\documentclass[conference]{IEEEtran}
\usepackage{blindtext, graphicx}

\usepackage{mathtools}
\usepackage{geometry}
\usepackage{balance}
\usepackage{graphicx}
\usepackage{dblfloatfix}
\usepackage{tabularx}
\usepackage{url}
\usepackage[caption=false,font=footnotesize]{subfig}

\usepackage{amssymb}
\usepackage{color}

\newcommand{\gr} {{\mathbb G}}
\newcommand{\op}[1] {{\overline{\mathcal P}}_{\gr}(#1)}
\newcommand{\ip}[1] {{\underline{\mathcal P}}_{\gr}(#1)}

\everymath{\displaystyle}

\begin{document}
\title{Efficient Retrieval of Logos Using Rough Set Reducts}

\author{\IEEEauthorblockN{Ushasi Chaudhuri}
\IEEEauthorblockA{Advanced Technology and\\Development Center,
IIT Kharagpur\\
Email: ushasi2cool@gmail.com
}
\and
\IEEEauthorblockN{Partha Bhowmick}
\IEEEauthorblockA{Computer Science and Engg Dept.\\
IIT Kharagpur\\
Email: pb@cse.iitkgp.ernet.in}
\and
\IEEEauthorblockN{Jayanta Mukherjee}
\IEEEauthorblockA{Computer Science and Engg Dept.\\
IIT Kharagpur\\
Email: jay@cse.iitkgp.ernet.in }}

\maketitle

\begin{abstract}
Searching for similar logos in the registered logo database is a very important and tedious task at the trademark office. Speed and accuracy are two aspects that one must attend to while developing a system for retrieval of logos. In this paper we propose a rough-set based method to quantify the structural information in a logo image that can be used to efficiently index an image. A logo is split into a number of polygons, and for each polygon we compute the tight upper and lower approximations based on the principles of rough set. This representation is used for forming feature vectors for retrieval of logos. Experimentation on standard data set shows the usefulness of the proposed technique. It is computationally efficient and also provides retrieval results at high accuracy. 
\end{abstract}

\IEEEpeerreviewmaketitle

\section{Introduction}
With a boom in the number of startup companies emerging every year, designing a new logo has become a necessity. Each pre-existing company usually registers its logo in the state trademark office. Whenever a new company is to register its logo, it has to go through the previously registered logo-database and see if any of them resembles the proposed logo, before getting an approval. Normally this entire process is carried out manually, and is quite a tedious method. As the data set for the trademarked logo increases, we need a faster and automated method which can provide us with all the top similar resembling logos to that of the chosen one. We propose a method for a fast and efficient retrieval of the logos in this paper. 

Existing techniques of CBIR (Content Based Image Retrieval) retrieve images based on some of their features like colour, texture, shape, bag of visual words (BoVW), spatial orientation, etc. Most researches that take place nowadays involve these kinds of feature sets and retrieval methods like \cite{Vassilieva2009,Liu2013,Jhanwar}. However, still no fixed, best set of features has been found so as to have uniformly good results on varied data sets. Also, CBIR has two major problems, which are difficult to be resolved \cite{Rajashekhar2005}. First, segmentation of an image to break it into meaningful parts based on the above low-level features is still a challenge. Then, there is a huge gap between the features and the semantic expressions embedded in an image \cite{Deb2004,Jain2006}.\\
Shih and Piramuthu \cite{Shih2015} have developed a technique of media retrieval by using rank SVM and then augmented the data set for a large-scale text retrieval in books. Gagaudakis and Rosin \cite{Rosin} used a histogram-based scheme, as it maintains a lower complexity and high efficiency. They avoided segmentation of regions, as they considered it unreliable, and instead used a variety of schemes like incorporating shape feature and Delaunay triangulation as a performance enhancer over color-labelled histograms.

In 2012, Bai et al.\cite{Bai2012} developed an algorithm to fuse different similarity measures for robust shape retrieval through a semi-supervised learning framework, named co-transduction. Given two similarity measures and a query shape, their algorithm retrieves the most similar shape and assigns them to do a re-ranking. Fudos and Palios \cite{Fudos2000} also proposed a shape-based approach for image retrieval by gradually ``fattening" the query shape until a best match is found. The accuracy is comparatively less and has a poly-logarithmic time behaviour (for uniform distribution of shape vertices). A distance space densifying technique for shape and image retrieval was proposed by Yang et al. \cite{Yang2013}. They presented an approach wherein they add synthetic ghost points in distance space directly. Using these ghost points they achieved an improved accuracy in shape, as well as image retrieval. A few other notable works in this area include logo retrieval using cascaded sparse color-localized matching by Pandey et al. \cite{Pandey2014}, using a tree-based shape descriptor for a scalable logo detection by Wan et al. \cite{Wan2013}, and a modal analysis approach for the search of a reliable shape by Acqua and Gamba \cite{Acqua1998}. Swain and Ballard \cite{Swain1991} used a technique of identifying objects in an image using colour histogram. This method became very popular since it was robust to changes in object's orientation, scale, and viewing position.\\
For the retrieval of logos, Jain \cite{JAIN1996} addressed some of their features of the issues like speed, accuracy and stability, based on colour and shapes of the images. Rajshekhar et al. \cite{Rajashekhar2005} proposed a logo retrieval algorithm, which was based on extracting shape features by calculating the morphological pattern spectrum for different structuring elements. In this work, we propose an efficient method for binary logo retrieval using rough-set attributes. We create a reduct using these attributes and then retrieve images similar to the query images by using inverse Hough transform method \cite{Duda1972}. For a faster searching algorithm, we use a $k$-d tree \cite{Bentley}. The proposed algorithm significantly gains in the average run time, as shown in Sec. 5.

\begin{figure}
\includegraphics[scale=0.15]{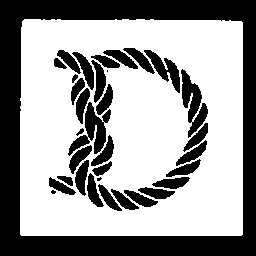}
\includegraphics[scale=0.15]{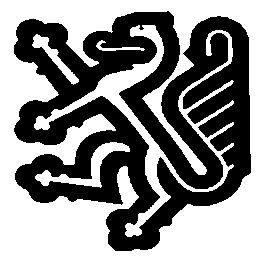}
\includegraphics[scale=0.15]{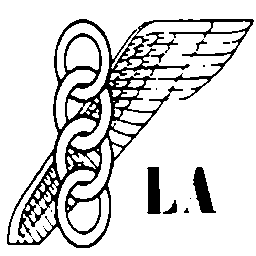}
\includegraphics[scale=0.15]{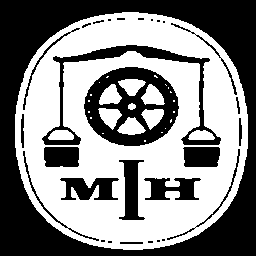}
\includegraphics[scale=0.15]{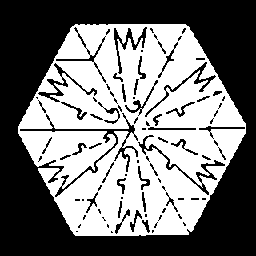}
\includegraphics[scale=0.15]{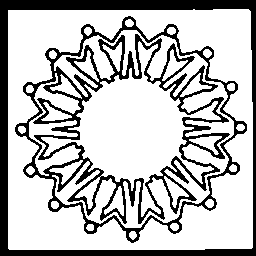}
\includegraphics[scale=0.15]{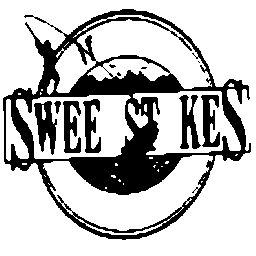}
\includegraphics[scale=0.15]{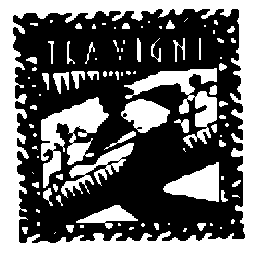}
\includegraphics[scale=0.15]{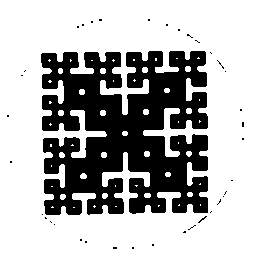}
\includegraphics[scale=0.15]{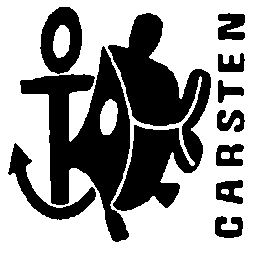}
\includegraphics[scale=0.15]{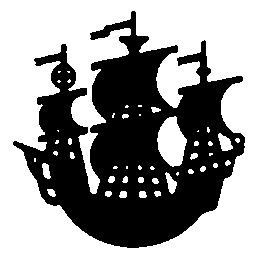}
\includegraphics[scale=0.15]{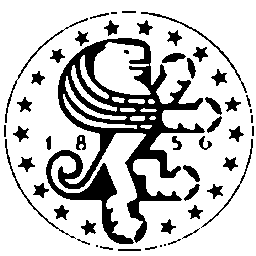}
\includegraphics[scale=0.15]{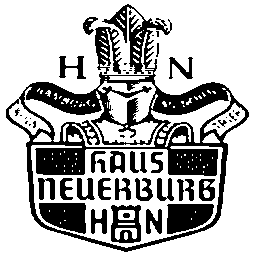}
\includegraphics[scale=0.15]{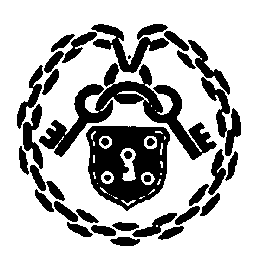}
\includegraphics[scale=0.15]{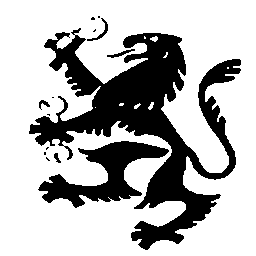}
\includegraphics[scale=0.15]{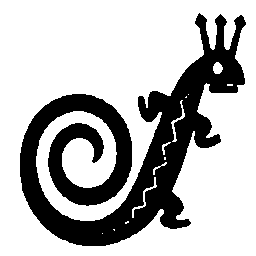}
\includegraphics[scale=0.15]{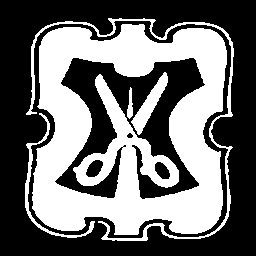}
\includegraphics[scale=0.15]{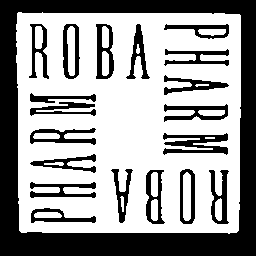}
\includegraphics[scale=0.15]{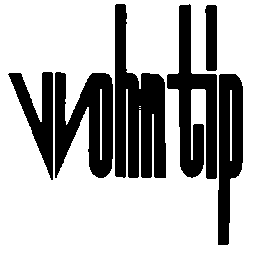}
\includegraphics[scale=0.15]{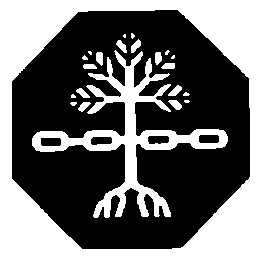}
\caption{Sample logo images from the data set.}\label{sample}
\end{figure}

For a 2D digital object $S$ and a grid $\gr$, we construct $\op{S}$ and $\ip{S}$, the tight upper approximation and tight lower approximations, respectively.

\section{Rough Set Reduct}
Rough sets, as defined by Pawlak \cite{Pawlak82,Pawlak91}, have been used in two stages. First, we create approximated lower and upper boundaries of the objects and then use them for approximation of the attributes for construction of the reduct. For constructing the boundary approximations, we use the concepts discussed in \cite{Biswas}. Since images in the data set are of the same dimensions, we use a fixed grid size of 3 for all the logo images. To construct the reduct \cite{Bag}, the rough-set approximation of each attribute and necessary details are discussed next.

\subsection{Black-to-White Ratio }
\begin{figure}[h!]
\centering
\subfloat[BW=0.25]{\includegraphics[scale=0.17]{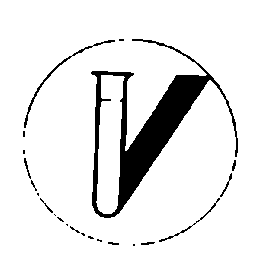}}
\subfloat[BW=0.5]{ \includegraphics[scale=0.16]{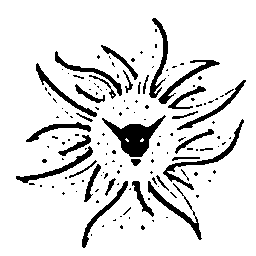}}
\subfloat[BW=1]{ \includegraphics[scale=0.16]{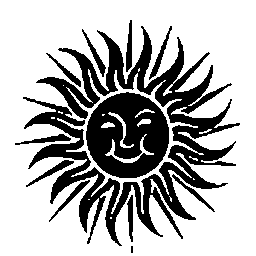}}
\subfloat[BW=2]{ \includegraphics[scale=0.16]{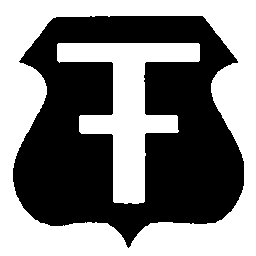}}
\subfloat[BW=4]{ \includegraphics[scale=0.16]{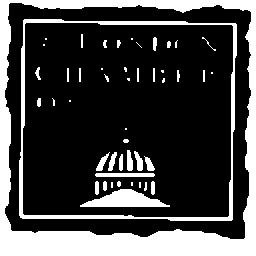}}
\caption{Illustration of black-to-white ratio attribute.}\label{1c}
\end{figure}

While traversing along the boundary of the object for the construction of the rough-set polygon, we also keep a count of the number of black and of the white pixels. This black-to-white pixel ratio (BW) gives us an approximate idea about the composition of the image. We quantize this ratio into 5 bins as 0.25, 0.5, 1, 2, and 4. Fig.~\ref{1c} illustrates a set of five examples.

\subsection{Number of Polygons and Major Polygons}
\begin{figure}[h!]
  \centering
\subfloat[TP=35 \& MP=10]{\includegraphics[scale=0.5]{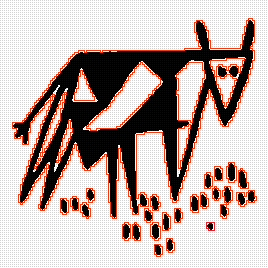}}
\subfloat[TP=41 \& MP=10]{ \includegraphics[scale=0.5]{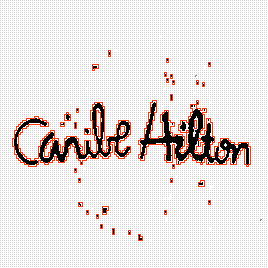}}
 \caption{Total number of polygons and major polygons.}\label{1d}
 \end{figure}

Another important primary feature is the number of polygons generated for a logo. This includes the parent polygons as well as the hole polygons. This feature gives a count of the absolute number of polygons. However it is seen that for some cases, the absolute number of polygons is not a good measure (Fig.~\ref{1d}). Further, the number of polygons can vary over different image orientations. Hence, we take account the variance in the size of polygons, and take into consideration only the major polygons (MP). Fig~\ref{1d} shows that due to the high variance in size of the smaller outer polygons to that of the bigger polygons, they are ignored. Fig.~\ref{1d}(a) contains 35 total polygons (TP), while only 10 are major polygons. Fig.~\ref{1d}(b) contains 41 total polygons, but only 10 are major polygons.

\subsection{Euler Number }
From various grid polygons that we have obtained from distinct components of a logo image (refer Fig. \ref{Fig:hole}), we observe that some components cover the entire region inside the polygon (primary polygon), while some have voids inside them. These voids, if they are big enough, also get represented by a separate grid polygon. These void polygons are referred to as hole polygons. To distinguish a primary polygon from a hole polygon, we notice that the difference in the number of $90^{\circ}$ vertices and $270^{\circ}$ vertices is 4 for the primary polygons. That is, the number of  $\measuredangle$ 90$^{\circ}$ $-$ the number of $\measuredangle$ 270$^{\circ}$ = 4. Similarly, for a hole polygon the number of  $\measuredangle$ 90$^{\circ}$ $-$ the number of $\measuredangle$ 270$^{\circ}$ = $-$4. Usually a primary polygon contains none or many hole polygons inside, represented by Euler Number (EN). To get this information, we define it as $2 - n$, where $n$ is the total number of polygons in  $\op{S}$.
In Fig.~\ref{Fig:hole}, we have two primary polygons, 1 and 3. Here, $n$ = 2 for polygon 1, and $n$ = 5 for polygon 3, whereby EN = 0 and $-3$, respectively for them. The hole polygons do not have any Euler number associated with and we use INV (invalid) to their values for demonstration. Both the hole polygon and the parent polygon have different containment relations, which are explained below. 

\subsubsection{Hole Containment (HC)}
Each hole polygon is enclosed within a primary polygon. We provide polygons with identification number in order to have the details of their containment relationship. This detail helps us to calculate the Euler number for each primary polygon. Fig.~\ref{Fig:hole} shows that the hole polygons 4, 5, 6, and 7 are contained within the polygon 3, while the hole polygon 2 is contained within the parent polygon 1.

\subsubsection{Polygon Containment (PC)}
Similar to the hole containment relation, each primary polygon is sometimes contained in another polygon. The parent polygon of a primary polygon gives us relevant information about the relative containment of each polygon within them.  Fig.~\ref{Fig:hole} illustrates the primary polygon containment relations. The primary polygon 3 is contained within the primary polygon 1. 

\subsection{Relative Hole Position}
\begin{figure}[h!]
  \centering
\subfloat[Holes=5; Parent=2]{ \includegraphics[scale=0.55]{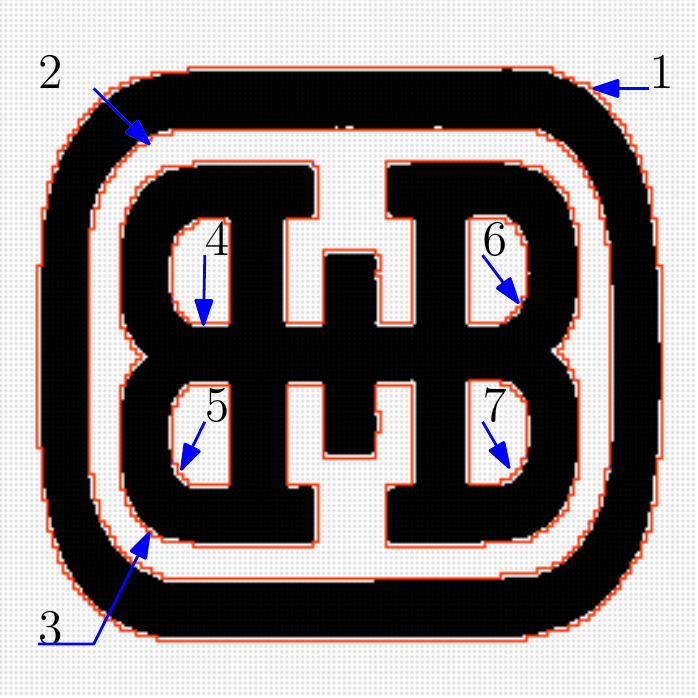}}
 \caption{Demonstration of Euler number and containment relations. }\label{Fig:hole}
  \vspace{-8pt}
 \end{figure}

Once we get the Euler number, we find the relative positions of holes (PoH). The hole centroid is found in the local coordinate with the top-left vertex $v_o$ of the polygon as the reference point. The relative position of each hole polygon's centroid $c$ is found with respect to $v_0$ of the outer primary polygon, i.e., $\op{S}$.
As a rough-set approach, we denote the position of hole (PoH) attribute: $c$ of polygons 4 and 5 lies to the right of $v_0$ of polygon 3 in Fig. \ref{Fig:hole}, and the $c$ of polygons 6 and 7 lies to the left of $v_0$ of polygon 3. We assign `$-$' and
`$+$' for the left and right lateral halves with respect to the point $v_0$ and `1' and `2' for the upper and lower halves. In Fig.~\ref{Fig:hole}, we observe how the logos are differentiated by their hole positions.

\subsection{Concavity}
 The rough-set polygon is constructed by three types of vertices, 90$^{\circ}$, 270$^{\circ}$, and 180$^{\circ}$. These vertices are defined as Type 1, $-1$, and 0, respectively. Concavity act as a very important characteristic to define the shape of a grid polygon \cite{Bag1,Biswas1}. A concavity is defined as the occurrence of at least two consecutive Type $\langle-1, -1\rangle$ vertices. If we get more than two consecutive Type $-1$ vertices, then it is a nested concavity. We classify the concavities into 4 types depending upon its orientation, namely left (L), right (R), upward (U), and downward (D). By preserving the order of these concavities, we estimate the structural similarities between two polygons using string edit distance. However, with a varying grid size, these concavities might get affected. Hence, we use an empirically chosen grid size of 3 in our experiments.

\subsection{Edge Ratio}
A grid polygon is comprised of horizontal and vertical line segments of fundamental unit length of $\gr$. So, while traversing on the list of perimeter components, we get a rough idea about the vertical to the horizontal perimeter component. From this, we calculate the summation of the horizontal perimeter component (HPC) and the summation of the vertical perimeter 
component (VPC). For each polygon in $\op{S}$, we define horizontal perimeter component (HPC) as the sum of lengths of 
its horizontal edges and vertical perimeter component (VPC) as that corresponding to its vertical edges. The ratio VPC:HPC is called edge ratio (ER). It is discretized to the nearest value in $\{\frac12, 1, 2\}$. Fig.~\ref{Fig:vdc} shows the significance of this feature with respect to the sample logo images.

\subsection{Directional Changes }
While traversing along the boundary of the rough set cover of an object, starting from the top-left vertex, the number of times we encounter a change in vertical direction (i.e., d=1 to d=3 or the vice versa), we get a change in vertical direction. These vertical changes can also be visualized as vertical `U-turns'. The number of times we encounter a U-turn, we increase our vertical direction change (VDC) count by 1. In case of horizontal directional change (HDC), instead of the top-left vertex, we start from the leftmost vertex and count the number of directional changes. Each U-turn is defined by a vertex sequence where two consecutive vertices are of Type $\langle +1, +1\rangle$ or $\langle -1, -1\rangle$. 

\begin{figure}[h!]
  \centering
  \subfloat[VDC=6; HDC=12; ER=0.5]{\includegraphics[scale=0.4]{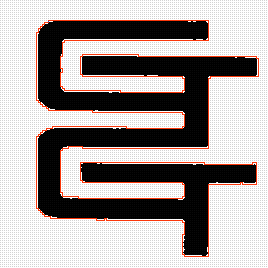}}
  \subfloat[VDC=14; HDC=14; ER=2]{\includegraphics[scale=0.4]{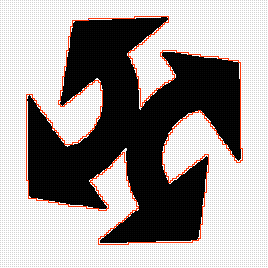}}
 \caption{Example showing the difference in Edge Ratios and direction changes.}\label{Fig:vdc}
  \vspace{-5pt}
 \end{figure}

In Fig.~\ref{Fig:vdc} we show an example of two images having a single polygon to demonstrate the values of VDC and HDC in them.
 
 \begin{figure}[h!]
  \centering
 {\includegraphics[scale=0.55]{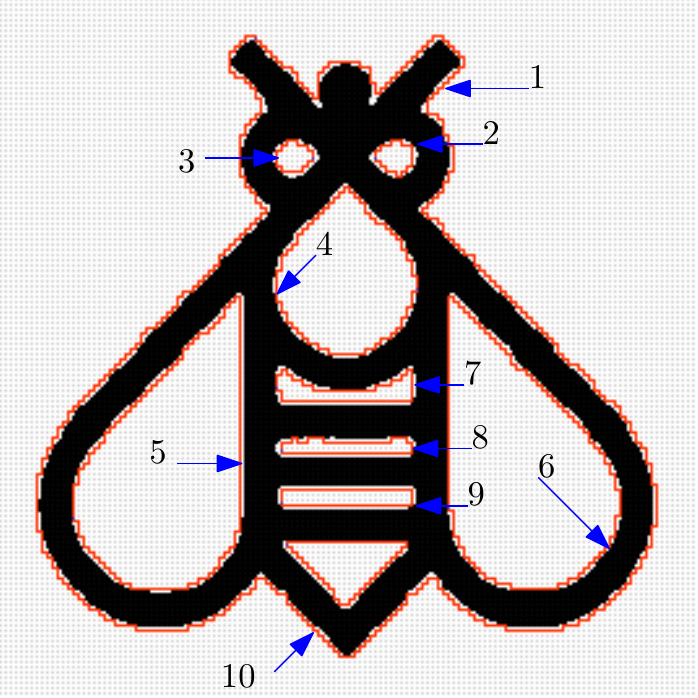}}
 \caption{TP=10; MP=10; Holes=9; Parents=1; BW=0.55}\label{refer}
 \vspace{-5pt}
 \end{figure}

\section{Retrieval using Inverse Hough Transform}
In Table \ref{Tab:1}, we show a subset of the rough set attributes for Fig. \ref{refer} to show how a logo image has some global image attributes and some polygon-specific attributes. We have 5 global attributes of an image (number of polygons, major polygons, hole polygons, parent polygons, black-to-white ratio) and 7 polygon-specific attributes (Euler number, hole containment, parent containment, vertical direction change, horizontal direction change, edge ratio and position of holes).
Depending upon the number of polygons contained by a logo image, the attribute size varies. For example, an image having $n$ polygons will have $n \times 7$ attributes plus 5 image attribute. We see that these attributes are well-discernible from each other and justify their merit in our image retrieval system. To calculate the similarity of an image to a query image, we use Euclidean and string edit distance, whichever is applicable. For example, Eucledean distance is used for VDC and edit distance for concavity.
 
 \begin{table}[t]
\begin{center}
\caption{Sample information table (shown partially) for the logo image shown in Fig.~\ref{refer} containing the object properties against the reduct.}
\begin{tabular}{||c |c| c| c|c |c|c|c|}
\hline
No & EN &HC & PC&VDC & HDC& ER & PoH\\
\hline\hline
1 & $-8$ & 0& 0& 10 & 10 & 1 & 0\\
\hline
2 & inv & 1& 0& 3 & 2 & 1 & $+2$\\
\hline
3 & inv & 1& 0& 3 & 2 & 1 & $+2$\\
\hline
4 & inv & 1& 0& 3 & 2 & 1 & $+2$\\
\hline
5 & inv & 1& 0& 3 & 2 & 2 & $-2$\\
\hline
6 & inv & 1& 0& 3 & 2 & 2 & $+2$\\
\hline
7 & inv & 1& 0& 5 & 3 & $1/2$ & $+2$\\
\hline
8 & inv & 1& 0& 3 & 3 & $1/2$ & $+2$\\
\hline
9 & inv & 1& 0& 3 & 3 & $1/2$ & $+2$\\
\hline
10 & inv & 1& 0& 2 & 3 & 1 & $+2$\\
\hline
\end{tabular}
\label{Tab:1}
\end{center}
 \vspace*{-4mm}
\end{table}

The data set we use comprises 1034 binary logo images, which was used in \cite{JAIN1996}. The feature database is stored in a CSV file and whenever we get a query image, this database is searched upon for the top 5 matches. The main challenge here is that each image has a different polygon number and hence has a varied size of feature vector. So we cannot rely on the distance between any two images by simple distance measures. Hence, we have implemented an inverse Hough-transform-based method, invoking a $k$-d tree based search operation.

\begin{figure}[h!]
  \centering
 {\includegraphics[scale=0.5]{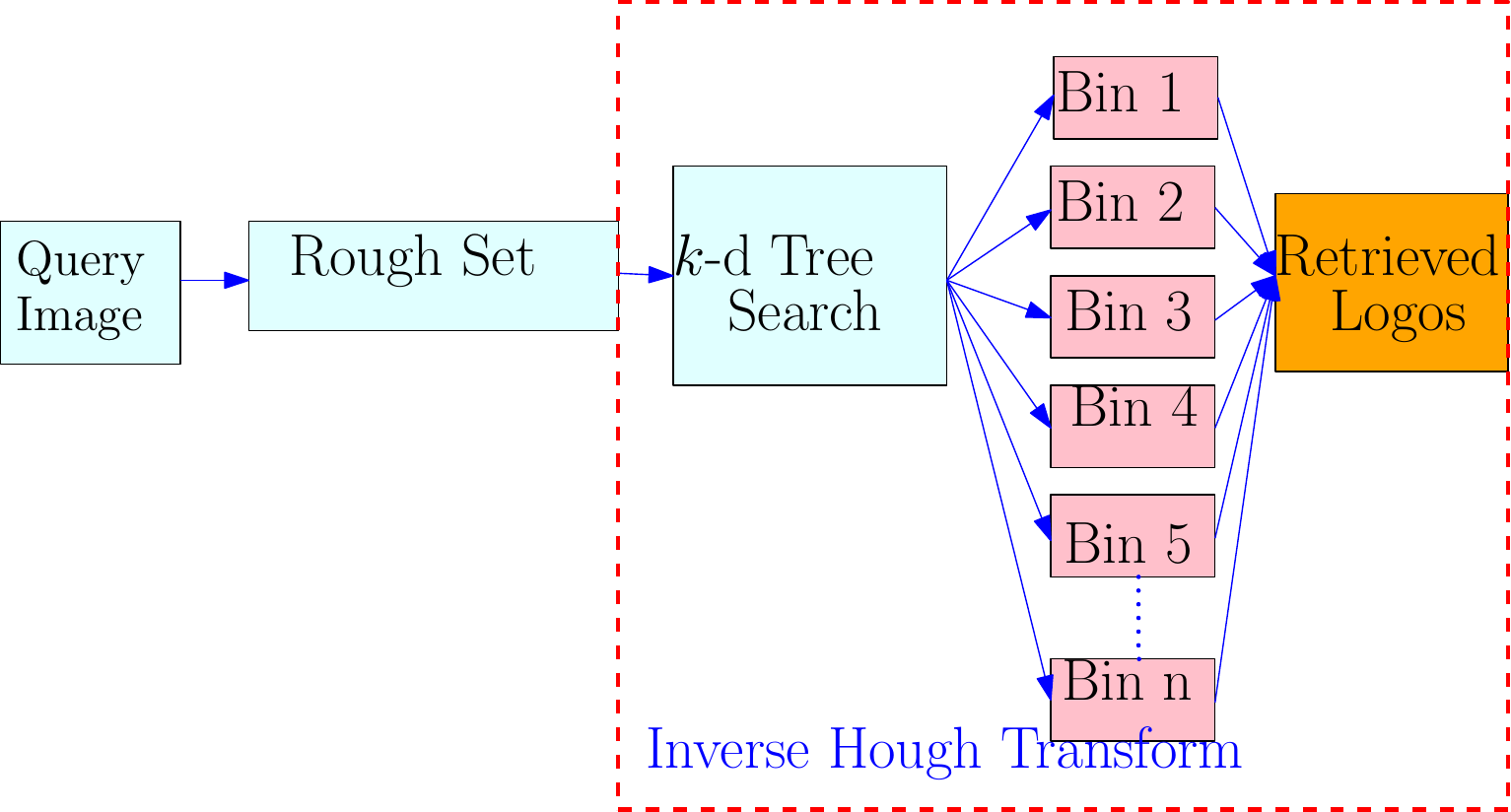}}
 \caption{Illustration of the proposed algorithm using rough set reduct}\label{prop}
 \vspace*{-1mm}
 \end{figure}

In the global feature list of each image, we set number of holes and number of parent polygons attributes as two dimensions in a $k$-d tree, and see that we get 306 distinct feature points. So, each feature point approximately contains 4 to 5 images having the same attributes binned under them. Whenever we get a query image, we find its attributes and search the point in this $k$-d tree in an $\mathcal{O}(\log{}n)$ time complexity. Once we get the point, we retrieve the image IDs and their features, stored in the form of a look-up table, and find the nearest 5 matches. For finding the matches, each polygon feature prototype is matched with each polygon from the image bin and if the distance is greater than a particular weighted Euclidean distance, then we add a counter to that image. The containment relations, Euler numbers, and the directional change attributes are found to be more robust and hence given more weights, compared to the remaining attributes. After each polygon-to-polygon matching, we find the image which has got the maximum counter and hence retrieve the nearest-match logos. Fig.~\ref{prop} shows the basic pipeline of our algorithm.

\section{Experimental Results}
The data set that we use comprises of 1034 challenging binary logo images \cite{JAIN1996}. Each image has a dimension of $256 \times 256$ pixels. A sample of the data set is already shown in Fig.~\ref{sample}. For testing the accuracy of our model, we synthetically performed degradation and transformation on the data sets. These operations include rotation, affine transform, addition of salt-and-pepper noise, erosion and dilation operations on test images. We used these images as query images and found the top 5 matches using the above discussed method. For retrieving the images, we need an average CPU time of 0.692 secs. This is computationally very efficient considering the size of the dataset of images. The CPU time is achieved with a machine of 64-bit Intel $\textregistered$ Core \textsuperscript{TM} i5-4210U processor, with 8GB RAM. We achieve a Mean Average Precision (MAP) value of 0.94 over 20 query images. The images that failed to get recognized are the ones with salt-and-pepper noise. These noise sometimes adds to the number of polygon and hence affects the precision. We can achieve a further improved model by de-noising the images.

\begin{figure*}[h]
  \centering
 {\includegraphics[scale=0.3]{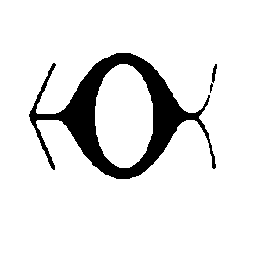}}
 {\includegraphics[scale=0.3]{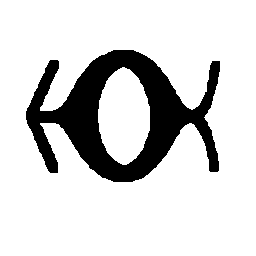}}
 {\includegraphics[scale=0.25]{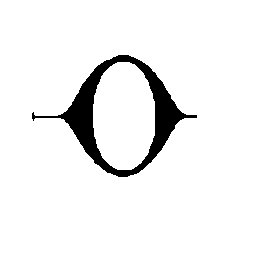}}
 {\includegraphics[scale=0.25]{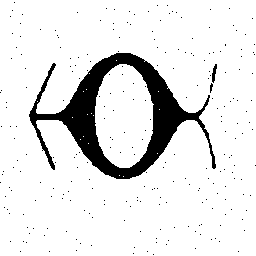}}
 {\includegraphics[scale=0.25]{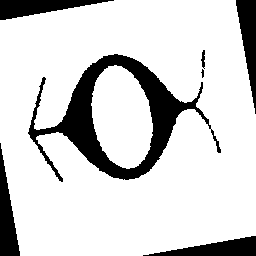}}
 {\includegraphics[scale=0.25]{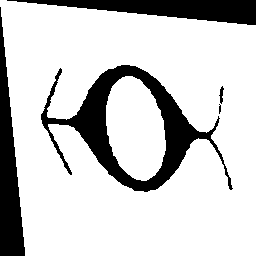}}\\ 
 {\includegraphics[scale=0.35]{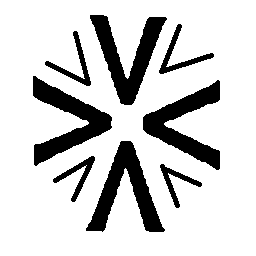}}
 {\includegraphics[scale=0.25]{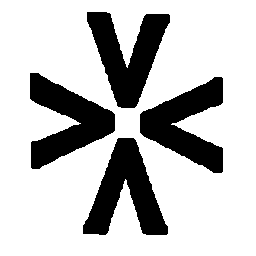}}
 {\includegraphics[scale=0.25]{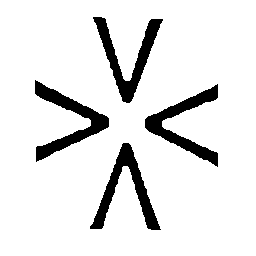}}
 {\includegraphics[scale=0.25]{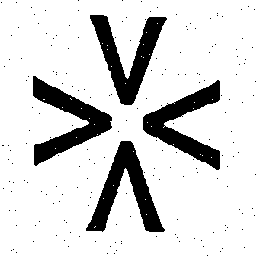}}
 {\includegraphics[scale=0.25]{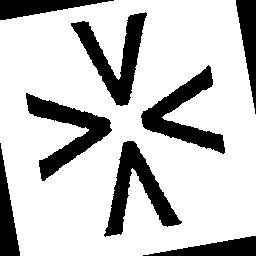}}
 {\includegraphics[scale=0.25]{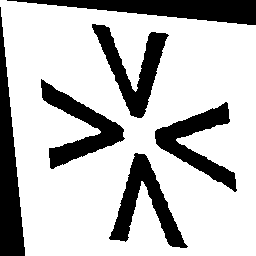}}\\
 {\includegraphics[scale=0.35]{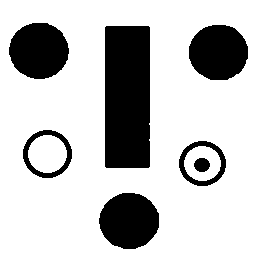}}
 {\includegraphics[scale=0.25]{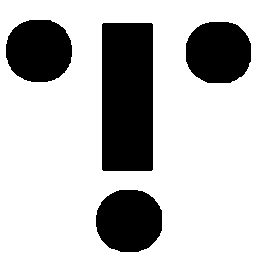}}
 {\includegraphics[scale=0.25]{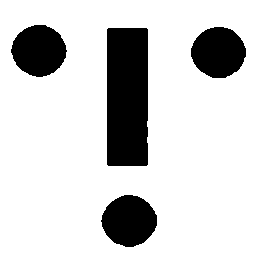}}
 {\includegraphics[scale=0.25]{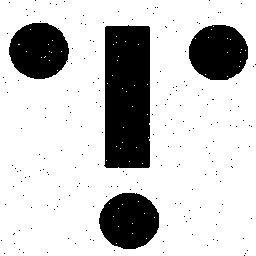}}
 {\includegraphics[scale=0.25]{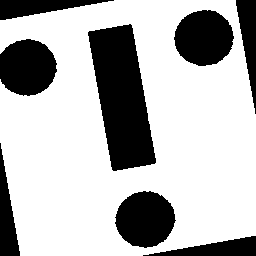}}
 {\includegraphics[scale=0.25]{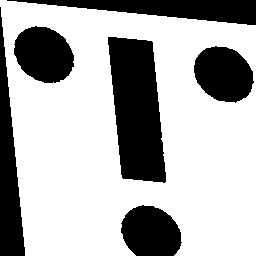}}
 \caption{Three sets of retrieved images shown in three rows. The first image is given as query.}\label{retrieved}

 \end{figure*}
 
\section{Conclusions}
We showed that a rough set model with a small-cardinality reduct can effectively retrieve logos very quickly and efficiently over a varied set of logos. A significant amount of reduction in the timing as well as an increase in the MAP have been found in this method compared to the existing methods. The reduct designed can be further used for various other discriminating and classifying areas like PDF reader, other Indian language reader and electrical symbol classifications.

\addcontentsline{toc}{section}{\numberline{}References}
\bibliographystyle{IEEEtran}
 \bibliography{references}

\end{document}